%% file: fedskel.tex
\documentclass[sigconf]{acmart}
\AtBeginDocument{%
  \providecommand\BibTeX{{%
    \normalfont B\kern-0.5em{\scshape i\kern-0.25em b}\kern-0.8em\TeX}}}

\copyrightyear{2021}
\acmYear{2021}
\setcopyright{acmcopyright}
\acmConference[CIKM '21]{Proceedings of the 30th ACM International Conference on Information and Knowledge Management}{November 1--5, 2021}{Virtual Event, QLD, Australia}
\acmBooktitle{Proceedings of the 30th ACM International Conference on Information and Knowledge Management (CIKM '21), November 1--5, 2021, Virtual Event, QLD, Australia}
\acmPrice{15.00}
\acmDOI{10.1145/3459637.3482107}
\acmISBN{978-1-4503-8446-9/21/11}

\usepackage{booktabs}
\usepackage{multirow}
\usepackage{algorithm}
\usepackage{algorithmicx}
\usepackage{algpseudocode}
\usepackage{subcaption}
\usepackage{adjustbox}
\usepackage{enumitem}
\usepackage{xcolor}
\usepackage{color}

\begin{document}
% \fancyhead{}
\title{FedSkel: Efficient Federated Learning on Heterogeneous Systems with Skeleton Gradients Update}


\author{Junyu Luo}
\email{luojunyu@buaa.edu.cn}
\affiliation{
    \institution{}
    \city{SCSE}
    \country{Beihang University}
}

\author{Jianlei Yang}
\authornote{Corresponding author is Jianlei Yang. Email: \href{mailto:jianlei@buaa.edu.cn}{jianlei@buaa.edu.cn}}
\email{jianlei@buaa.edu.cn}
\affiliation{%
    \institution{}
    \city{SCSE}
    \country{Beihang University}
}

\author{Xucheng Ye}
\email{yexucheng@kuaishou.com}
\affiliation{%
    \institution{Heterogeneous Computing Center,}
    \country{Kuaishou Technology}
}

\author{Xin Guo}
\email{guoxin@act.buaa.edu.cn}
\affiliation{
    \institution{}
    \city{SCSE}
    \country{Beihang University}
}

\author{Weisheng Zhao}
\email{weisheng.zhao@buaa.edu.cn}
\affiliation{
    \institution{}
    \city{SME}
    \country{Beihang University}
}

\thanks{This work is supported in part by the National Natural Science Foundation of China (Grant No. 62072019, 61602022), State Key Laboratory of Software Development Environment and the 111 Talent Program B16001.
The source code of this paper is publicly available on:
\url{https://github.com/BUAA-CI-Lab/FedSkel}.}


\input{Docs/0-abstract.tex}


\begin{CCSXML}
<ccs2012>
   <concept>
       <concept_id>10010147.10010178.10010219</concept_id>
       <concept_desc>Computing methodologies~Distributed artificial intelligence</concept_desc>
       <concept_significance>500</concept_significance>
       </concept>
   <concept>
       <concept_id>10010147.10010919</concept_id>
       <concept_desc>Computing methodologies~Distributed computing methodologies</concept_desc>
       <concept_significance>100</concept_significance>
       </concept>
   <concept>
       <concept_id>10003120.10003138</concept_id>
       <concept_desc>Human-centered computing~Ubiquitous and mobile computing</concept_desc>
       <concept_significance>300</concept_significance>
       </concept>
 </ccs2012>
\end{CCSXML}

\ccsdesc[500]{Computing methodologies~Distributed artificial intelligence}
\ccsdesc[100]{Computing methodologies~Distributed computing methodologies}
\ccsdesc[300]{Human-centered computing~Ubiquitous and mobile computing}

\keywords{Federated learning, heterogeneous system, distributed learning}

\maketitle

\input{Docs/1-introduction.tex}
\input{Docs/2-methodologies.tex}
\input{Docs/3-experiments.tex}
\input{Docs/4-conclusion.tex}

\bibliographystyle{ACM-Reference-Format}
\bibliography{acmart}
\end{document}

%% file: Docs/0-abstract.tex
\begin{abstract}

Federated learning aims to protect users' privacy while performing data analysis from different participants. However, it is challenging to guarantee the training efficiency on heterogeneous systems due to the various computational capabilities and communication bottlenecks. In this work, we propose \texttt{FedSkel} to enable computation-efficient and communication-efficient federated learning on edge devices by only updating the model's essential parts, named skeleton networks. \texttt{FedSkel} is evaluated on real edge devices with imbalanced datasets. Experimental results show that it could achieve up to $5.52 \times$ speedups for CONV layers' back-propagation, $1.82\times$ speedups for the whole training process, and reduce $64.8\%$ communication cost, with negligible accuracy loss. 

\end{abstract}

%% file: Docs/1-introduction.tex
\section{Introduction}

Federated learning~(FL)~\cite{konevcny2016federated} was proposed to protect data privacy while making use of the data collected from different participants. FL on edge devices or mobile devices utilizes multiple clients to collaboratively train the identical model on large amounts of local data. A global model can be trained by exchanging parameters between clients and servers instead of directly using private data. In practice, Google deployed FL to enhance models for emoji prediction~\cite{ramaswamy2019federated} and query suggestions~\cite{yang2018applied} applications.

Heterogeneities among clients will lead to inefficiency~\cite{lim2020federated}, which is a challenging issue when deploying FL systems. The heterogeneity caused by imbalanced computational capabilities~\cite{ignatov2019ai} and communication bandwidth may significantly degrade the training speed. The global training process is limited by the slower clients, i.e., stragglers. If faster clients wait for slower ones, the overall training speed will become slow. Otherwise, the data in slower clients cannot be well utilized for global model learning.

\begin{figure}[t]
    \centering
    \includegraphics[width=0.43\textwidth]{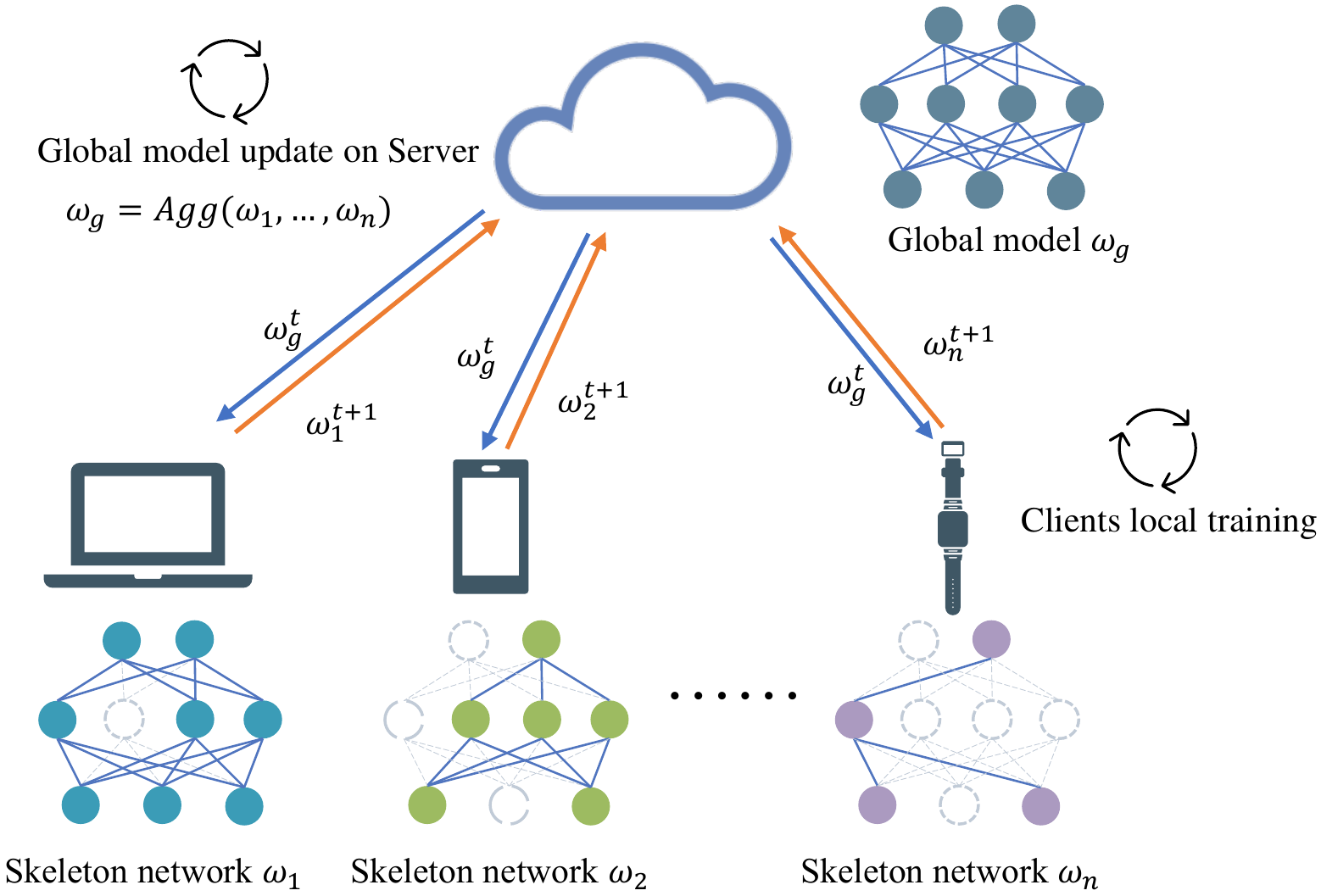}
    \caption{The framework of \texttt{FedSkel}. Each client selects its skeleton network and only updates (i.e. train, upload to server, download from server) the skeleton network.
    }
    \label{fig:fedskel-introduction}
\end{figure}

Statistical heterogeneity is another challenge. In FL, data distribution across clients is inherently Non-IID (non-identically independently distributed). Thus, it is difficult for the shared global model to generalize for all clients. Some previous works~\cite{liang2020think, li2020lotteryfl} tried to train personalized models on different clients. We also attempt to utilize data heterogeneity to make clients focus on different skeleton networks.

In this work, we propose \texttt{FedSkel} to improve FL efficiency on heterogeneous systems. As shown in Figure~\ref{fig:fedskel-introduction}, each client determines its personalized skeleton network and it will only train and up/download the skeleton network. By adjusting the size of clients' skeleton networks, \texttt{FedSkel} can reduce workloads on slower clients and balance the latency across different clients to achieve efficient FL systems.

The main contributions of this paper are as follows:
\begin{itemize}
\item We introduce a novel federated learning framework, \texttt{FedSkel}, which achieves better computation efficiency on single clients and communication efficiency on FL systems.

\item We propose to select personalized skeleton networks by a metric dynamically and perform gradients pruning to enable efficient training. We present the effectiveness of our method through experiments and analysis. 

\item  We implement and measure speedups of \texttt{FedSkel} on real devices. We also compare \texttt{FedSkel} with three baselines to demonstrate it will not degrade accuracy while accelerating FL systems.
\end{itemize}

\section{Related Works}

\textbf{Weight Pruning.} Weight pruning usually aims to accelerate models inference by reducing the model parameters after training. ~\cite{anwar2017structured, lebedev2016fast} exploit structural weight pruning. Yu~\cite{yu2018distilling} observed that some filters are more crucial than others on different categories of data. In this paper, we follow ~\cite{yu2018distilling} to select personalized skeleton networks for clients. Different from weight pruning methods, we mainly focus on reducing training workloads.\\
\textbf{Communication-Efficient Federated Learning.} Extensive prior works attempt to speed up FL by decreasing parameter communications~\cite{strom2015scalable, konevcny2016federated, shi2019distributed}. Lin~\cite{lin2017deep} demonstrated that removing redundant gradients will not hurt accuracy. However, these works did not take efforts to reduce the computation of local devices. \texttt{FedSkel} can not only achieve better communication efficiency but also accelerate whole training process.\\
\textbf{Personalization.} Personalization is a critical challenge in federated learning. LG-FedAvg~\cite{liang2020think} and LotteryFL~\cite{li2020lotteryfl} train personalized models for each client. However, these methods did not consider the heterogeneity of computational capabilities in real FL systems. In this paper, \texttt{FedSkel} can facilitate efficient FL training with personalization maintained.

%% file: Docs/2-methodologies.tex
\begin{figure}[t]
    \centering
    \includegraphics[width=0.37\textwidth]{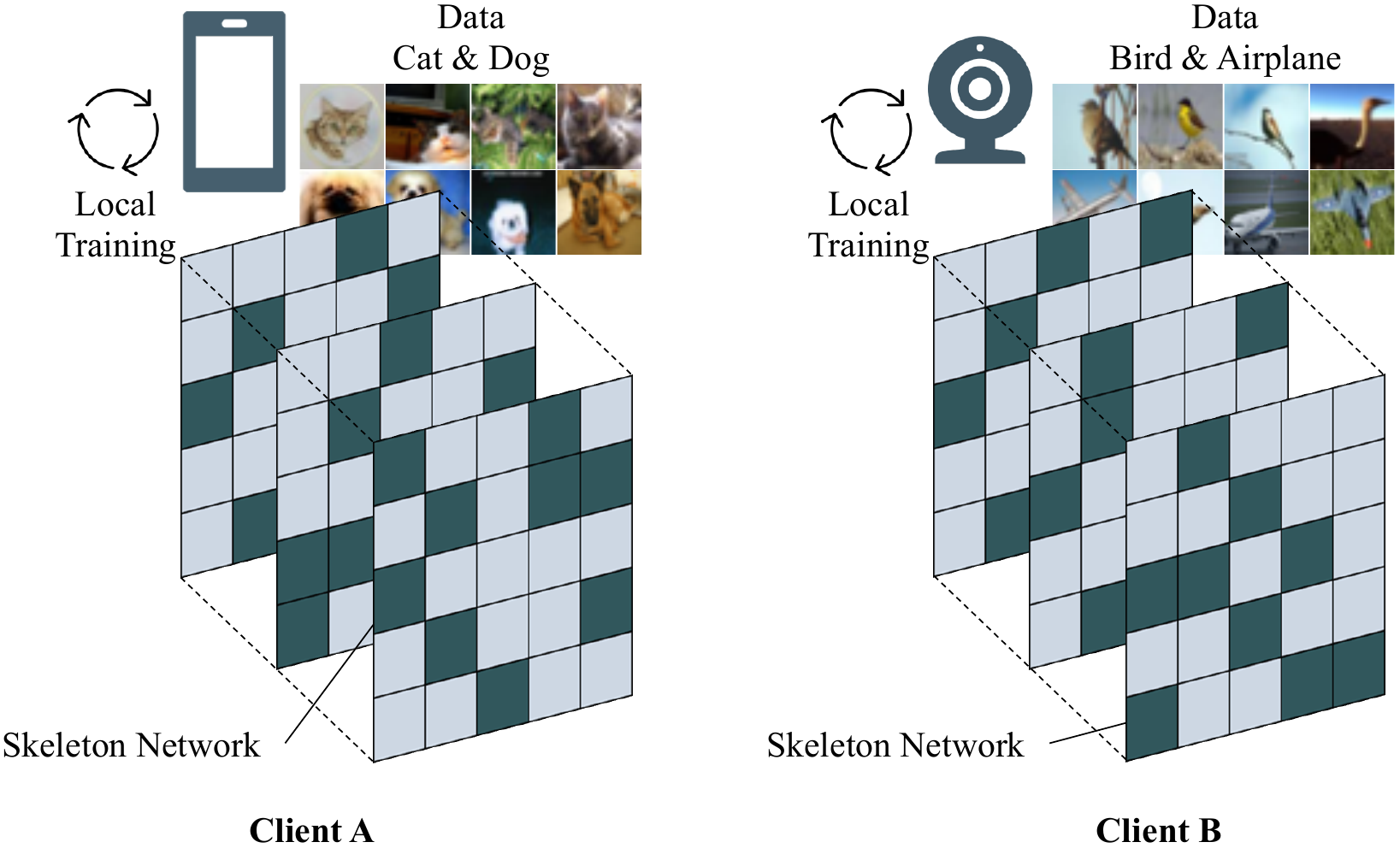}
    \caption{Different skeleton networks are existed in different clients since data distribution is imbalance in FL.}
    \label{fig-kernel-heatmap}
\end{figure}

\section{Methodologies}
\label{sec:method}

In FL, data imbalance is an inherent feature. Hence, submodels inherited from global model will be of different importance to different clients. From this motivation, we propose \texttt{FedSkel}. In \texttt{FedSkel}, each client only trains and up/downloads their important submodels, which are named skeleton networks. The central server aggregates all local skeleton networks to obtain the global model, which can handle tasks of different data distributions. Clients' training processes can be accelerated by only updating the skeleton networks.

\begin{figure}[t]
    \centering
    \includegraphics[width=0.27\textwidth]{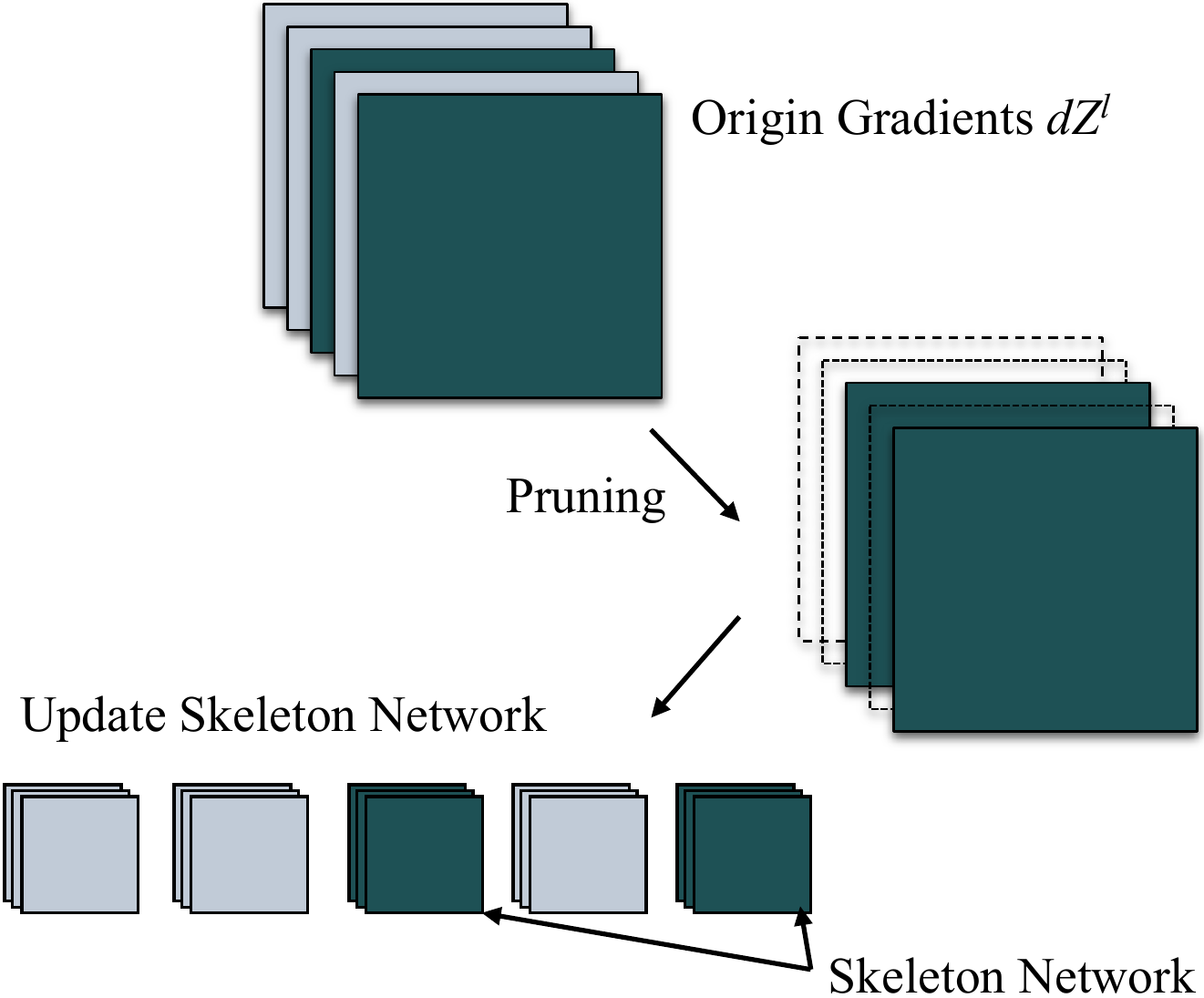}
    \caption{Structural gradients pruning for skeleton network update on clients training.}
    \label{fig:2-method-gradient-pruning}
\end{figure}

\subsection{Skeleton Network}
\label{sec:Skeleton-Network}
Yu~\cite{yu2018distilling} found there are category-related filters in CNNs, which are more activated to the specific data categories and contribute more to the prediction. In FL scenario, as shown in Figure~\ref{fig-kernel-heatmap}, clients' important filters~(skeleton network) are different since they hold data of different distributions. \\
\textbf{Importance Metric.}
Clients can select their skeleton network according to the following defined metric.
The logits of CNNs can be calculated with $l$-th layer by (notice that we left out bias here): 
\begin{equation}
\label{eq:CNN-Inference}
\centering
Z\left( W, A^l \right) = \left(\alpha... \alpha\left(A^{l}W^{l+1} \right)...W^{L-1}\right) W^L ,
\end{equation}
% and 
\begin{equation}
\label{eq:CNN-M1}
\centering
{M_i^l} = \left| A^l_i \right|,
\end{equation}
where the model is with total $L$ layers, $Z$ is the output, $W^l$ and $A^l$ is the weight and input to the $l$-th layer, $\alpha$ is the activation function.

As shown in Eq.~\ref{eq:CNN-Inference} and Eq.~\ref{eq:CNN-M1}, when $W^{l+1}$ is non-zero, ${M_i^l}$ determines the contribution of $A^l_i$ to logits. Hence, we adopt ${M_i^l}$ to measure the importance of $i$-th filter in $l$-th layer. \\
\textbf{Gradients Pruning enables Efficient Training.}
Gradients pruning is utilized to only train skeleton networks and adjust workloads.
CNNs training mainly includes three kinds of matrix multiplication operations, which take up most of the training computation~\cite{ye2019accelerating}:
\begin{itemize}
\item \emph{Forward}: ${{Z}^\textit{l} = {A}^{l-1} \ast {W}^l}$
\item \emph{Gradients Back-Propagation}: $\mathrm{d}{A}^{l-1}$ = $\mathrm{d}{Z}^l$ $\ast$ ${W}^l$
\item \emph{Weight Gradients Computation}: $\mathrm{d}{W}^l$ = ${A}^{l-1}$ $\ast$ $\mathrm{d}{Z}^l$
\end{itemize}
where $Z^l$ denotes the output of $l$-th layer.
The last two multiplication operations are involved in back-propagation procedure.
In this paper, we try to apply structured pruning on gradients $dZ^l$ as Figure~\ref{fig:2-method-gradient-pruning}. Since we compress the $dZ^l$, the computation in \textit{Gradients Back-Propagation} and \textit{Weight Gradients Computation} can be greatly reduced, thus reducing the workloads on clients.\\
\textbf{Gradients Pruning has Negligible Impact on Accuracy.} \cite{lin2017deep} revealed the redundancy of gradients. Ye~\cite{ye2019accelerating} exploited fine-grained gradient pruning for training acceleration. In this paper, structural gradients pruning is utilized since it is more hardware-friendly.
FedProx~\cite{li2018federated} demonstrated that adding constraints to the updates from clients could benefit the issue of data heterogeneity.
In \texttt{FedSkel}, we only update the submodels on clients, which also satisfy the constraints introduced by FedProx. Experiments in Section~\ref{sec:experiments} show that \texttt{FedSkel} will not degrade accuracy and can even perform better than baselines.

\subsection{\texttt{FedSkel}} \label{sec:fedskel}

\texttt{FedSkel} aims to solve the potential inefficiencies in heterogeneous FL systems. It tries to balance the workloads of different clients by setting different skeleton network ratios $r$.

The whole training procedure are divided into two alternately processes: \textit{SetSkel} and \textit{UpdateSkel}. In \textit{SetSkel} process, skeleton networks are determined for each client. In \textit{UpdateSkel} process, clients only update the skeleton network to reduce the training workloads. In practice, a \textit{SetSkel} process is usually followed by $3$ to $5$ \textit{UpdateSkel} processes.

\textbf{\underline{\textit{SetSkel} Process}}

The purpose of \textit{SetSkel} is to select the skeleton network for each client. At the same time, let clients get updates related to non-skeleton networks by exchanging parameters with the server.

\textbf{Server sets skeleton ratios $r$. } In a $n$-clients system, the $i$-th client uploads its computational capability $c_i$ to server. The server normalizes $c$ as $c'_i=c_i/c_{max}$. We simply try to set skeleton ratios $r$ with a linear function, and setting $r$ more effectively can be further explored. 

\begin{figure}[t]
    \centering
    \begin{subfigure}{0.9\linewidth}
         \includegraphics[width=0.9\textwidth]{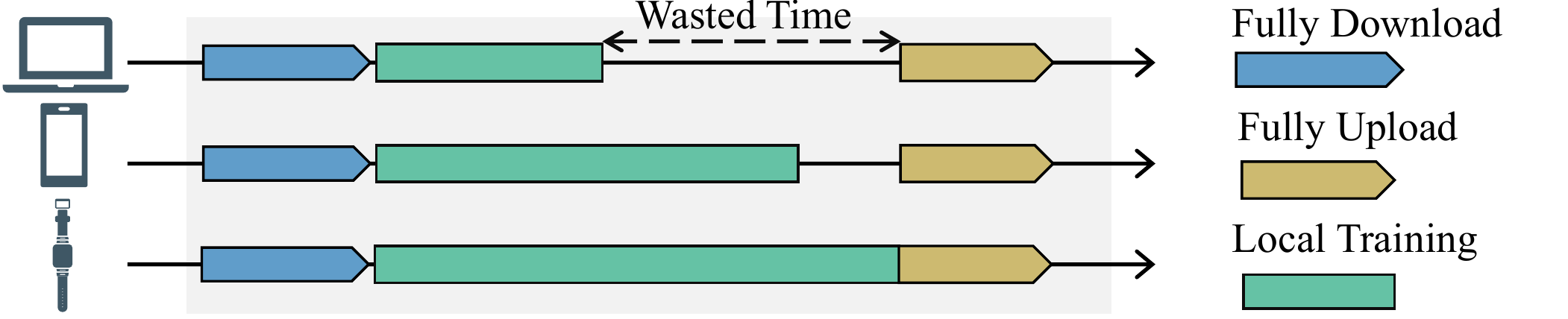}
        \caption{Original FL training process.} \label{fig:2-method-origin-process}
    \end{subfigure}
    \par
    \begin{subfigure}{0.9\linewidth}
        \includegraphics[width=0.9\textwidth]{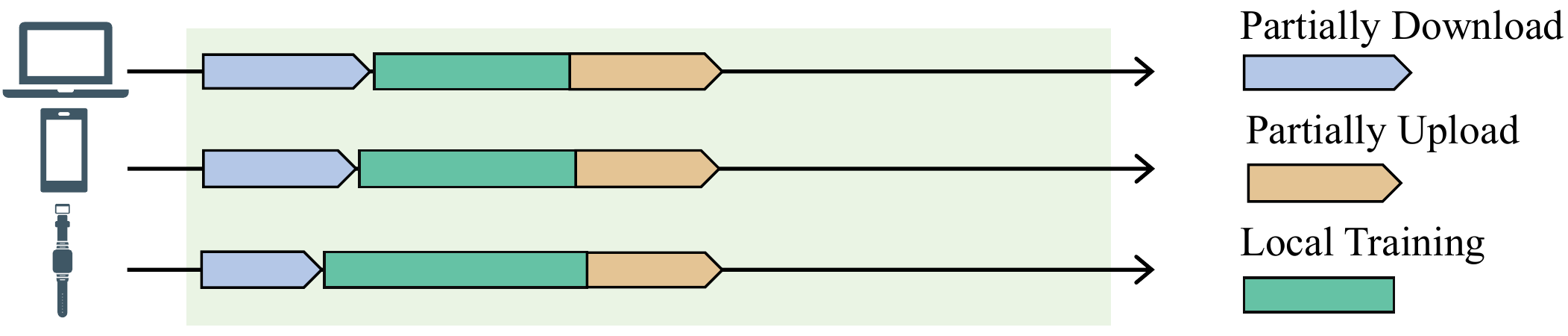}
        \caption{\textit{UpdateSkel} process in \texttt{FedSkel}.} 
        \label{fig:2-method-updateskel-process}
    \end{subfigure}
    \caption{\texttt{FedSkel} reduces both local training cost and parameters communication cost. Partially up/download means \texttt{FedSkel} only needs to up/download skeleton networks.}
    \label{fig:fedskel-flow-chart}
\end{figure}

\textbf{Calculating importance metric during training. } The \textit{SetSkel} process is similar to the standard FL process (as shown in Figure~\ref{fig:2-method-origin-process}). The only difference is that we accumulate the importance metric $M_i^l$ and set the skeleton network according to them.

\textbf{\underline{\textit{UpdateSkel} Process}}

In \textit{UpdateSkel} process, as shown in Figure~\ref{fig:2-method-updateskel-process}, clients reduce the computational workloads and communication workloads by only training and exchanging skeleton networks.

\textbf{Computation reduction. } In \textit{UpdateSkel} Process, clients only train the skeleton network, the computational workloads can be reduced as described in Section~\ref{sec:Skeleton-Network}.

\textbf{Communication reduction. } In \textit{UpdateSkel} Process, clients only up/download parameters on skeleton network to/from the server. The amount of communication cost is determined by skeleton network ratio $r$.

\textbf{\underline{Overall Procedure}}

\textit{SetSkel} processes and \textit{UpdateSkel} processes are iterated. In real world, \textit{SetSkel} processes typically conduct in the period when computational resources are idle (such as in nights), so they do not take up lots of computational resources. The server adopts federated averaging~\cite{mcmahan2016communication} to aggregate updates to obtain the global model which can generalize to all type of data distribution. 

Hence, \texttt{FedSkel} improves the learning efficiency by reducing both the computational workloads and communication cost.

%% file: Docs/3-experiments.tex
\section{Experiments}
\label{sec:experiments}
In this section, we demonstrate that \texttt{FedSkel} is more training-efficient on heterogeneous systems without affecting accuracy. 
  
\begin{table}[t]
\centering
\caption{Speedups on Intel CPU and ARM CPU with different skeleton ratio $r$.}
\begin{adjustbox}{width=0.35\textwidth}
\begin{tabular}{ccccc}
\toprule
\multirow{2}{*}{$r$} & \multicolumn{2}{c}{Intel CPU} & \multicolumn{2}{c}{ARM CPU}             \\
\cmidrule[0.05em](lr){2-3}
\cmidrule[0.05em](lr){4-5}
& Back-prop & Overall & Back-prop & Overall \\ 
\hline
$40\%$ & $2.08\times$ & $1.10\times$ & $1.94\times$ & $1.35\times$ \\
$30\%$ & $2.57\times$ & $1.13\times$ & $3.06\times$ & $1.52\times$\\ 
$20\%$ & $3.38\times$ & $1.21\times$ & $4.32\times$ & $1.61\times$\\ 
$10\%$ & $5.52\times$ & $1.28\times$ & $4.56\times$ & $1.82 \times$\\ 
\bottomrule
\end{tabular}
\end{adjustbox}
\label{tab:speedup-eval}
\end{table}

\begin{figure}[t]
    \centering
    \includegraphics[width=0.33\textwidth]{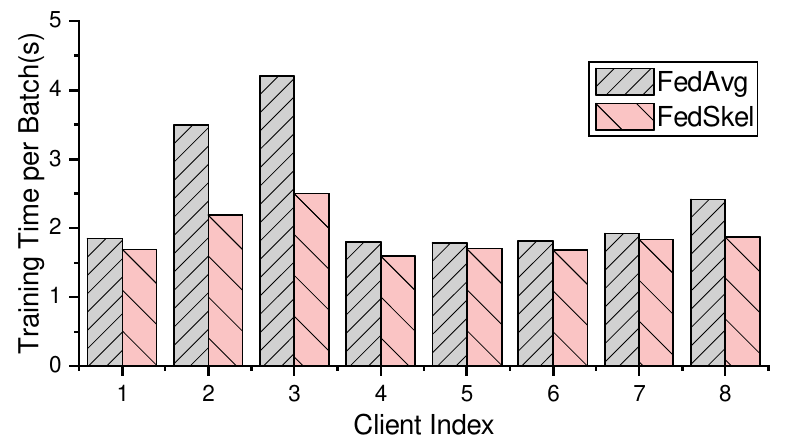}
    \caption{Runtime of each client for an 8-device system to train one batch with \texttt{FedSkel} and FedAvg, including forward and backward.}
    \label{fig:experiment-accelerate-edge-systems}
\end{figure}
  
\subsection{Acceleration}
\label{sec:speedup-experiments}
In this part, we evaluate the speedups of \texttt{FedSkel} in \textit{UpdateSkel} processes. We conduct experiments on a single device and a system of 8 devices, respectively.

\textbf{Implementation.}
\texttt{FedSkel} is evaluated on Intel Xeon E5-2680 v4@2.4GHz CPU with MKL and Raspberry Pi 3B+ (ARM v8@1.4GHz) with OpenBLAS.
We modify the back-propagation in Caffe's~\cite{jia2014caffe} CONV layer.
Speedups are measured with LeNet~\cite{lecun1998gradient} on MNIST~\cite{lecun1998gradient}.
Same as Section~\ref{sec:fedskel}, $r$ denotes the skeleton network ratio.

\textbf{Acceleration on Single Device.} To evaluate the performance with different $r$, we adopt $r = \{40\%,30\%,20\%,10\%\}$.
According to the results in Table~\ref{tab:speedup-eval}, we can achieve up to $5.52\times$ speedups at back-propagation of CONV layers and $1.82\times$ at whole training process.

\textbf{Acceleration on Edge Systems.} We also deploy experiments on a real system with $8$ Raspberry Pi.
Devices are set to different computational capabilities $c_i$ as in a heterogeneous system. We set $r_i$ according to $c_i$.
Figure~\ref{fig:experiment-accelerate-edge-systems} shows the time consuming in \textit{UpdateSkel} process for one batch (batch\_size=512) training.
It shows that \texttt{FedSkel} can balance the workloads, speedup the slower clients and achieve up to $1.82\times$ speedups of the whole system.

\subsection{Communication Cost Reduction}
\label{sec:experiments-communication}
\texttt{FedSkel} can significantly reduce communication cost since we only exchange updates on the skeleton network in \textit{UpdateSkel}. As shown in Table~\ref{table:comm-computation-compare}, \texttt{FedSkel} with $r=10\%$ can reduce $64.8\%$ communicate cost to the whole training process, including \textit{SetSkel} and \textit{UpdateSkel}. 

\begin{table}[t]
\centering
\caption{Volume of parameters communication for different methods with LeNet-5 on MNIST.}
\begin{adjustbox}{width=0.35\textwidth}
\begin{tabular}{ccc}
\toprule
Method  & Params Comm. & Reduction \\ 
\cmidrule[0.05em](lr){1-3}
FedAvg~\cite{mcmahan2016communication} & $12.8\times 10^9$ & $-$ \\
FedMTL~\cite{fedmtl} & $12.0\times 10^9$ & $6.3$\% \\
LG-FedAvg~\cite{liang2020think} &  $8.5\times 10^9$ & $33.6$\%   \\
\texttt{FedSkel} ($r=10\%$) & $\mathbf{4.5\times 10^9}$ & $\mathbf{64.8\%}$  \\ \bottomrule
\end{tabular}
\end{adjustbox}
\label{table:comm-computation-compare}
\end{table}

\subsection{Convergence and Accuracy}
\label{sec:convergence-accuracy}
In this part, we empirically compare \texttt{FedSkel} with the baseline methods. We demonstrate that \texttt{FedSkel} can significantly accelerate training without hurting accuracy.

\textbf{\underline{Experimental Settings}}
% \subsubsection{\underline{Experimental Settings}}

\textbf{Datasets and Models.}
All methods are evaluated on MNIST~\cite{lecun1998gradient}, FEMNIST~\cite{caldas2018leaf}, CIFAR-\{10,100\}~\cite{krizhevsky2009learning} datasets and with LeNet-5~\cite{lecun1998gradient} and ResNet-\{18,34\}~\cite{Resnet}.
We adopt the Non-IID data setting as ~\cite{liang2020think} did.
Each client is assigned with $2$ shards of Non-IID splited data for MNIST and CIFAR-10, while $20$ shards for others.
LeNet is trained for $1000$ epochs and ResNets for $600$ epochs.
Each \textit{SetSkel} process is followed by 3 \textit{UpdateSkel} processes.
To make a fair comparison, we follow the experimental design in ~\cite{mcmahan2016communication} and also exploit local representation learning. The FL system consists of $1$ central server and $100$ clients.
All methods are evaluated with the same settings.

\textbf{Heterogeneous System Settings.} In the real world, clients in FL systems are of different computational capabilities. To verify the performance of \texttt{FedSkel} in this scenario, we set each client with a different ratio $r$ equidistant ranging from $10\%$ to $100\%$. 

\begin{table}[t]
\centering
\caption{Accuracy comparison of baselines and \texttt{FedSkel} (ours) on different datasets with LeNet.}
\label{TestAcc-dataset}
\begin{adjustbox}{width=0.46\textwidth}
\begin{tabular}{@{}cccccc@{}}
\toprule
\multirow{2}{*}{Method}  & \multirow{2}{*}{Test Type\footnotemark[2]}& \multicolumn{4}{c}{Dataset} \\
\cmidrule[0.05em](lr){3-6}
 & & MNIST & FEMNIST & CIFAR-10 & CIFAR-100 \\
\hline % \cmidrule[0.05em](lr){1-6}
\multirow{2}{*}{FedAvg~\cite{mcmahan2016communication}}
& New & $99.09$ & $57.52$ & $\mathbf{59.03}$ & $32.44$  \\
& Local & $99.09$ & $57.52$ & $59.03$  & $32.44$ \\
\hline % \cmidrule[0.05em](lr){1-6}
\multirow{2}{*}{FedMTL~\cite{fedmtl}}
& New & $39.76$ & $24.69$ & $10.32$ & $2.15$ \\
& Local & $99.41$ & $29.93$ & $90.49$ & $42.68$ \\
\hline % \cmidrule[0.05em](lr){1-6}
\multirow{2}{*}{LG-FedAvg~\cite{liang2020think}}
& New & $99.09$ & $57.57$ & $58.48$ & $32.44$ \\
& Local & $99.45$ & $78.92$ & $92.47$ & $52.95$ \\
\hline % \cmidrule[0.05em](lr){1-6}
\multirow{2}{*}{\texttt{FedSkel}}
& New & $\mathbf{99.09}$ & $\mathbf{59.43}$ & $58.48$ & $\mathbf{32.52}$ \\
& Local & $\mathbf{99.46}$ & $\mathbf{82.95}$ & $\mathbf{92.60}$ & $\mathbf{53.66}$ \\
\bottomrule
\end{tabular}
\end{adjustbox}
\end{table}

\footnotetext[2]{We follow the test settings of LG-FedAvg~\cite{liang2020think}. Local Test (new predictions on an existing device), where we test each client and the clients' train/test data are of the same distribution. New Test (new predictions on new devices), where we test the global model on test data with the same distribution of the whole dataset.}

\begin{table}[t]
\centering
\caption{Accuracy comparison of baselines and \texttt{FedSkel} (ours) with different models on CIFAR-10 dataset.}
\begin{adjustbox}{width=0.4\textwidth}
\begin{tabular}{ccccc}
\toprule
\multirow{2}{*}{Method} & \multirow{2}{*}{Test Type\footnotemark[2]} & \multicolumn{3}{c}{Model} \\
\cmidrule[0.05em](lr){3-5}
% \cline{3-5}
& & LeNet & ResNet-18 & ResNet-34  \\
\hline

\multirow{2}{*}{FedAvg~\cite{mcmahan2016communication}} 
& New & $\mathbf{59.03}$ & $67.61$ & $70.28$   \\
& Local & $59.03$ & $67.61$ & $70.28$  \\ 
% \cmidrule[0.05em](lr){1-5}
\hline

\multirow{2}{*}{FedMTL~\cite{fedmtl}} 
& New & $10.32$ & $10.92$ & $10.02$  \\
& Local & $90.49$ & $92.78$ & $91.90$ \\
% \cmidrule[0.05em](lr){1-5}
\hline

\multirow{2}{*}{LG-FedAvg~\cite{liang2020think}} 
& New & $58.48$ & $75.67$ & $\mathbf{76.95}$  \\
& Local & $92.47$ & $96.21$ & $97.34$ \\
% \cmidrule[0.05em](lr){1-5}
\hline

\multirow{2}{*}{\texttt{FedSkel}} 
& New & $58.48$ & $\mathbf{77.20}$ & $76.92$ \\
& Local & $\mathbf{92.60}$ & $\mathbf{96.59}$ & $\mathbf{97.65}$ \\
 
\bottomrule
\end{tabular}
\end{adjustbox}
\label{TestAcc-models}
\end{table}

\textbf{\underline{Experimental Results}}

As shown in Table~\ref{TestAcc-dataset} and Table~\ref{TestAcc-models}, crossing different datasets and models, our method will not hurt accuracy while achieving acceleration. \texttt{FedSkel} can even improve the personalization to achieve better accuracy on the local test.

\subsection{Analysis and Discussions}

\texttt{FedSkel} will not affect accuracy though it prunes gradients in the training process. Several reasons are accounting for it.
\begin{itemize}[leftmargin=*]
\item Skeleton network is the submodel which has a more crucial impact on predict results. Hence the combination of skeleton network is able to perform well on each task.
\item \texttt{FedSkel} facilitates personalization by only updating skeleton networks. It enables clients to perform better on their own tasks.
\item  According to FedProx~\cite{li2018federated}, fewer updates to the global model facilitate robust convergence. \texttt{FedSkel} only updates the skeleton network, also contributes to a faster and stable convergence.
\end{itemize}
Experiments show that FL is robust to gradients pruning. By optimizing the gradients' flow, we can achieve an efficient and personalized FL system. 

%% file: Docs/4-conclusion.tex
\section{Conclusions}
In this work, we propose \texttt{FedSkel} as a new framework for heterogeneous FL systems. In our method, clients select skeleton networks and only update skeleton filters. Skeleton network ratios are adaptive to clients' computational capabilities. We have shown that our method does not affect the accuracy through extensive experiments and analysis, and achieves up to $5.52\times$ speedups in back-prop and $1.82\times$ speedups in the overall process on edge devices. Our approach enables efficient FL on heterogeneous systems.
The future work can be the better metrics of selecting skeleton networks, strategies to set clients' skeleton ratios.
We will also extend our explorations on the effect of gradient optimization in FL systems.